# Predicting resonant properties of plasmonic structures by deep learning

Iman Sajedian[1], Jeonghyun Kim[1] and Junsuk Rho[1,2,*]

[1] Department of Mechanical Engineering, Pohang University of Science and Technology (POSTECH), Pohang 37673, Republic of Korea

[2] Department of Chemical Engineering, Pohang University of Science and Technology (POSTECH), Pohang 37673, Republic of Korea

* Corresponding author email: jsrho@postech.ac.kr

**Abstract:** Deep learning can be used to extract meaningful results from images. In this paper, we used convolutional neural networks combined with recurrent neural networks on images of plasmonic structures and extract absorption data form them. To provide the required data for the model we did 100,000 simulations with similar setups and random structures. By designing a deep network we could find a model that could predict the absorption of any structure with similar setup. We used convolutional neural networks to get the spatial information from the images and we used recurrent neural networks to help the model find the relationship between the spatial information obtained from convolutional neural network model. With this design we could reach a very low loss in predicting the absorption compared to the results obtained from numerical simulation in a very short time.

## 1. Introduction

Novel machine learning methods can find solutions for complex problems. They have shown their power in image classification, identify objects in images, finding labels for images, text translation or voice recognition as a few examples. It is not far from mind that they can be generalized to help in other fields of science too. For example they have been used in high-energy physics for finding exotic particles[1], or in Biology for predicting the sequence specificities of DNA-and RNA [2]. In here, we are going to use deep learning to extract optical information from images of plasmonic structures. The idea introduced here can be generalized to use in other fields of optics.

Deep learning and other machine learning methods' goal is to find a pattern in a given data. They can learn from and make predictions on data. As a simple machine learning method we can name the curve fitting method. In curve fitting we can fit a curve on a given data, and by that curve we can predict new information in new coordinates. In the same manner, we will fit a model on some given images and predict information from new images. In order to do that, we need a much more advanced machine learning method compared to curve fitting. Deep learning is a subfield of machine learning family which refers to methods for image processing, voice recognition, natural language processing, etc. that can help us achieve this goal.

Here, we are going to use two machine learning methods combined together for extracting optical information from given structures. We are going to use convolutional neural network (CNN) which is famous for image classification [3-7] to extract spatial information from images (like lines, curves, edges, their orientation,…). And recurrent neural Network (RNN) which is used for time series problems [8], or voice recognition [9] to help the model find the relation between spatial information Fig. 1 .

By combining these two models we will be able to find the absorption curve of plasmonic structures, by which we can find resonant frequencies [10]. Resonant frequencies are the frequencies in which the absorption reaches its maximum value. Other optical properties can be found in the same way as we will discuss later. Resonant properties of plasmonic structures have applications in many fields like sensors [11, 12], waveguides [13, 14] or in photovoltaic devices [15] and nonlinear optics [16, 17] as a few examples.

## 2. Theory and implementation

Convolutional neural networks are consisted from layers which are designed to extract spatial information from images. They extract information from sub parts of images and use them as new input data for other layers. In convolutional neural networks layers are usually consisted from input layer, convolutional layers, pooling layers, dropout layers, fully connected layers and output. Each of these layers can or cannot be used, based on the structure of the problem [18, 19]. On the other hand are recurrent neural networks that are used to find the relation between the input data. For example they can find different meanings of a same word based on its position in a sentence.  RNNs have different kinds of layers. In here, we used gated recurrent unit (GRU) layer [20] which found to be more efficient than long short-term memory (LSTM) [21] which is another famous RNN layer.

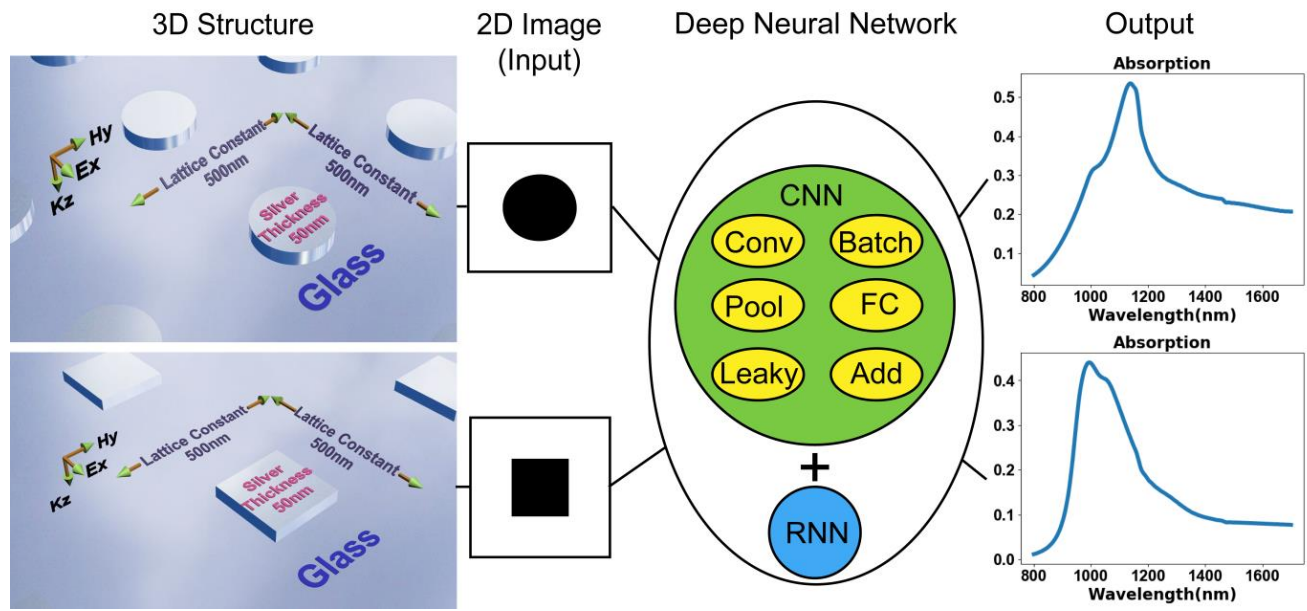

Fig. 1: A deep neural network can predict the absorption curve from the 2D image of a given structure. 3D structure with similar properties and different shapes can be modeled as 2D images. All the geometrical properties like lattice constant, material type, polarization and thickness should be the same. Under these conditions the shape of the structure can be shown as a 2D image. This image is fed into the deep neural network. The network can predict the absorption curve from the given images.

We divide our discussion into three parts. First, we discuss how we prepared the data for the input of the deep learning model. Then we describe the model layout that we used and its parameters and its implementation. And in the last part we will compare the results obtained from the model with the unseen data obtained from FDTD simulations.

**2.1 Input data**

As mentioned before our goal is to extract optical information (absorption curves in here) from the image of a given structure. We first discuss how we can convert a 3D physical structure to a 2D image. We know that if a 3D structure doesn't have variations in one dimension we can simulate it in 2D by taking a snapshot of its cross section. For example an infinite cylinder in 3D can be simulated as a circle in 2D.  In the method that we are introducing here we can convert another group of 3D structures to 2D structures which is only applicable in machine learning methods. In machine learning methods we can omit any variable that does not have variance, which means it is constant for all the input data. For example assume that all the structures have the same thickness of 50nm, this means that we can omit thickness from our input data. The obvious advantage is that the objects don't need to be infinite in order to be simulated as 2D objects. To illustrate the idea more let's get back to the problem of curve fitting. If you add a constant to all the input data, the shape of the final curve won't change. It is just that you know that every new predicted point that you find by this new curve is added by that fixed constant you added at the first place. The same happens here. If we keep some of the geometrical properties of the structure as constants (for the input and the output) our model will work for all the cases which have the same geometrical properties.

## Varaiance In The Input And Output

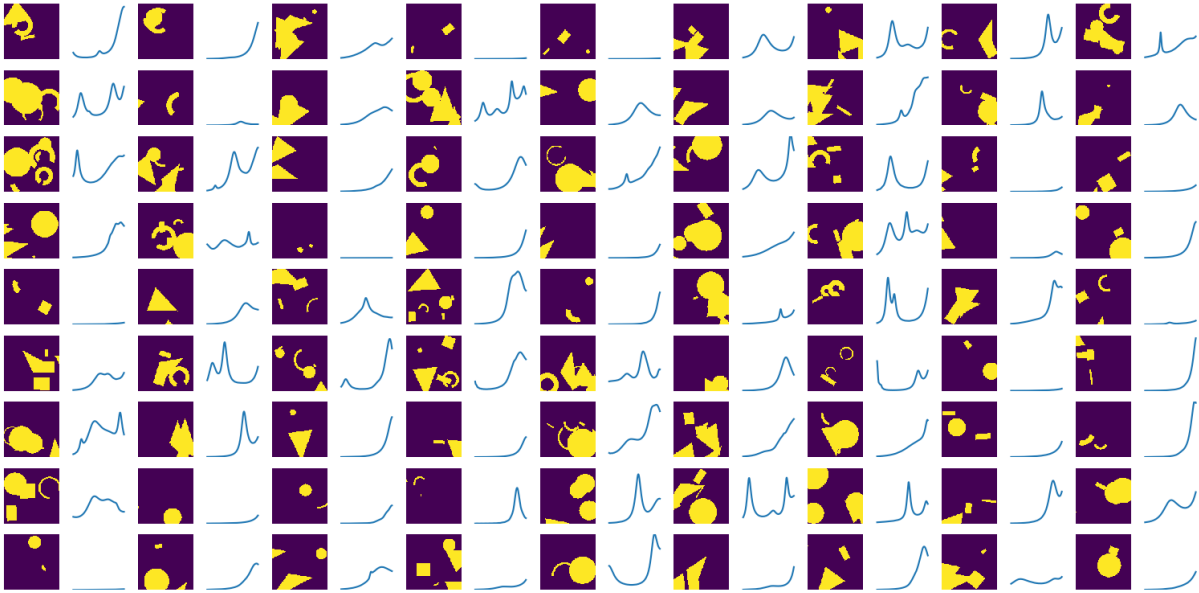

Fig. 2: A small number of images used as the input of the model and their corresponding absorption curves. Total number of 100,000 structures were simulated. Yellow areas mean silver and violet areas mean empty space (for better visualization we used yellow and violet instead of black and white). We tried to prepare enough images to cover the variance needed for the model by using random geometrical parameters. The curves are absorption for wavelengths from 800nm to 1700nm. We removed the axes so we can show more structures.

The fixed geometrical properties are the material type of the substrate (glass), the lattice constant (500nm), the polarization of the source (as shown in fig 1.), the boundary conditions (periodic boundary conditions), the thickness of the structure (50nm) and the material type of the sample (silver). We selected these geometrical properties randomly just to show how the idea works. Using these conditions helped us to convert our 3d structure to a 2D black and white image as can be seen in fig 1. Please beware that by fixing these parameters we did not change the physics of the problem. We still maintained all the geometrical information needed. When we fix the mentioned geometrical properties the model no longer needs to know them (since they are the same for all structures) and will work with any given 2D images. The fact that we are using only one material type will help us to use black and white images for the input. If we wanted to use more material types we should use color images which leads to 3 added additional channels for the input data(for Red, Green and Blue). Using black and white picture means that wherever there is silver in our structure we have black pixels in the input image and wherever there is no silver we have white pixels in the input image. The resolution that we used for images was 100×100 pixels. These numbers were selected for two reasons. First we picked a resolution high enough so that we don't lose any detail form the structure (the higher the resolution, more details can be covered). And second for not running into memory problems when trying to compile the model (the higher the resolution means the higher input data and so the possibility of memory shortage).

To ensure a high accuracy for the model we prepared 100,000 simulations, which means that we prepared 100,000 images of different structures and 100,000 absorption curves. Some of the structures and their corresponding absorption curves are shown in fig 2.

The input data should have enough variance so that the model can predict any new given structure. This variance should be either in the geometrical part or in the physical part (the absorption curves in our problem). Let's clarify this with an example. Imagine an image classification problem in which we want to classify cats and dogs from a number of animal images. Now there is two points that we should be cautious about. First we cannot classify all species of cats from the images of just for example Persian cats. If we feed our model with just Persian cat's images (no variance in the input), it cannot classify Bombay cats for us. So we should feed the network with the images of enough different types of cats so the model can learn how to recognize a cat from the input images. Second, imagine that we have given the output of the model in the training stage to be "cat" or "not a cat" (no variance in the output). This model can classify cats, but cannot classify dogs, because the required variance ("cat" and "dog") was not fed to the network's output too. It can just tell us if an image is a cat or not. The same thing happens in our model. Imagine that the absorption curves fed to the network are only consisted from absorption curves with no resonance or with just resonances in a small range of frequencies (like 800nm to 1000nm instead of 800nm to 1700nm). We cannot expect this model to predict resonance in all the frequencies needed. To overcome these obstacle the choice of the input shapes are very important.

To prepare the input data we used Lumerical scripting language. By using for loops we could create enough input data for our model. We used random values for the following parameters to create the random structures:

- Number of shapes in each structures (was chosen randomly from 1 to 6 shapes)
- Shapes' types (was chosen randomly from: circle, triangle, rectangle, ring, polygon)
- Position of each shapes (x and y position was set randomly)
- Shapes sizes (for example radius for circle, width and length for rectangle... Was chosen randomly)
- Rotation of each shape

With the help of above guideline we could provide as much data as we need and it had the required variance. A sample of prepared structures and their corresponding absorption curves are brought in Fig. 2. The absorption curves are obtained by using 1000 frequency points in Lumerical. So the output of our model should have 1000 nodes.

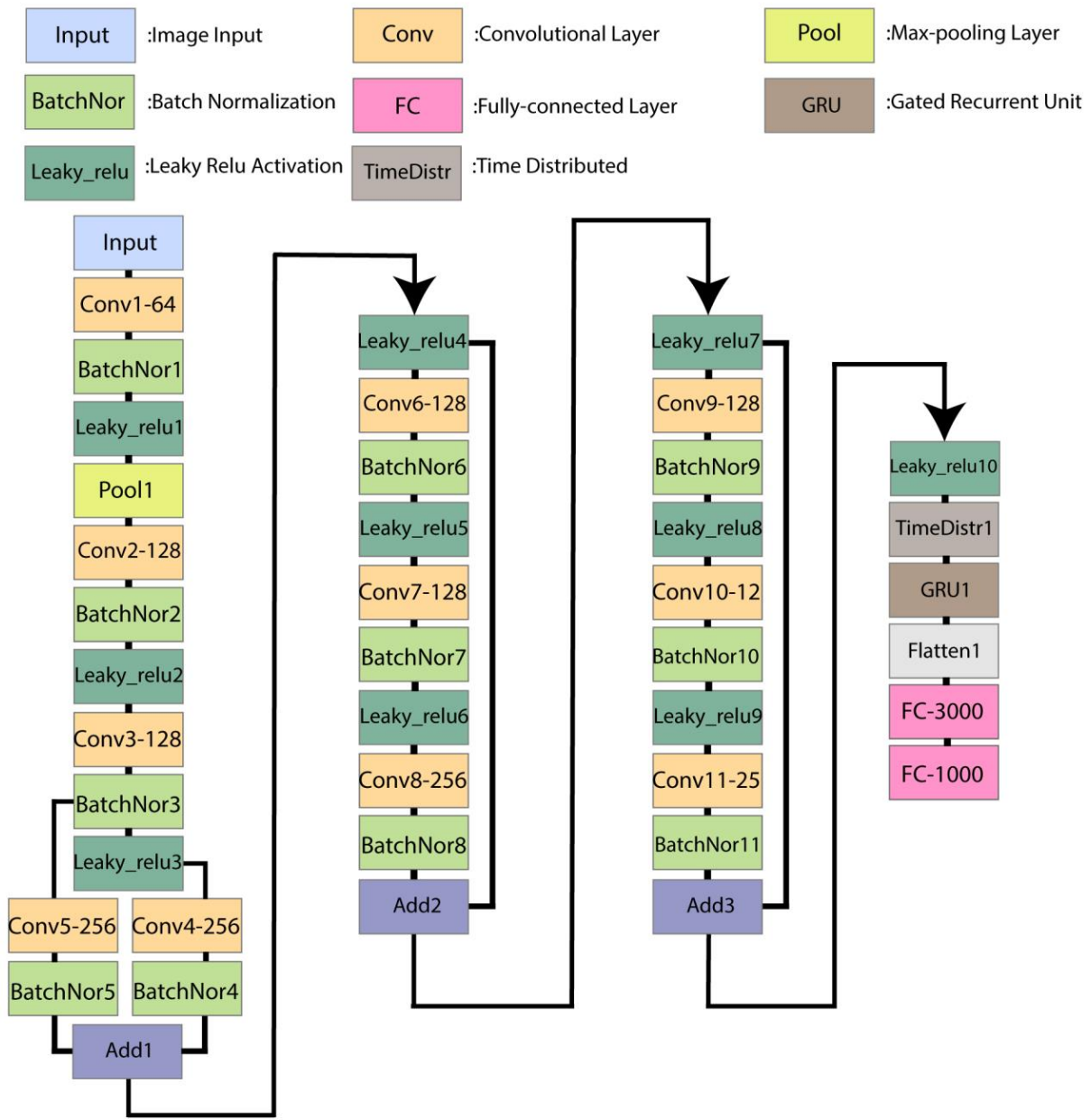

Fig. 3: The deep learning model layout. This number in the boxes shows the number of filters used in each layer. Total number of 500 epochs were run in the training stage, with a learning rate of 0.0001 and Nestrov Adam as optimizer. This model is a combination of ResNet CNN model and an RNN model. With this layout we could reach a very low loss on our test data. The final fully connected layer with 1000 nodes is our output, which each node is a frequency point in the absorption curve.

**2.2 Model layout and its implementation**

Once the input data is prepared, the next step is to find the right model for the problem. We have a large number of options ahead of us to choose, like number of layers, the number of nodes in each layer, the layers layout, the activation functions, the loss function and the optimizer and their hyper parameters. To find the correct combination of these options, we should try different combination of them to reach the lowest loss possible. We start by splitting our data into three parts. The first part is used to train the model, which is called the train dataset. The second part is used to test the model which is called the test dataset. And the last part is the validation dataset for validating the model. To assess the model's efficiency we change model's parameters like number of layers or some other parameter and each time that we change a parameter we check our model on the test dataset. In the end the best model is the one with the lowest loss on the test dataset. Once we find the best model that fits our data, we check the model on the validation dataset. The difference between the test dataset and the validation dataset is the fact that the test dataset is used to check different parameters to find the best model and the validation data is used to assess the final model. This stage assure us that the model works on the unseen data.

The splitting ratios that we used for the data was 60, 30, and 10. Which means that 60 percent of whole data was used for training, 30 percent was used for testing, and 10 percent was used for validation. The model parameters that we tuned, were number of convolution layers, number of nodes in each convolution layer, shape of the used strides in convolution layers, the model layout, number of neurons in fully connected layers, and the optimizer type. It takes a lot of practicing to find the right model, and to minimize the final loss as much as possible. The final layout that we reached for our model is shown in fig 3. It was a combination of residual network CNN known as ResNet[22] and a small recurrent neural network. ResNet architecture has this property that we can design a very deep network without getting a very small gradient [22] or running into memory problems. This is done by defining a shortcut connection which is resulted from the addition of a layer with one or more of its next layers.

The loss function that we used was mean squared error[23]:

$$M.S.E. = \frac{1}{n} \sum_{i=1}^{n} (Y_i - P_i)^2 \qquad (1)$$

In the above equation Y is the vector of real values and P is the vector of predictions. Mean squared error is always positive and values closer to zero mean higher accuracy. To avoid overfitting we used batch normalization and weight regularization.

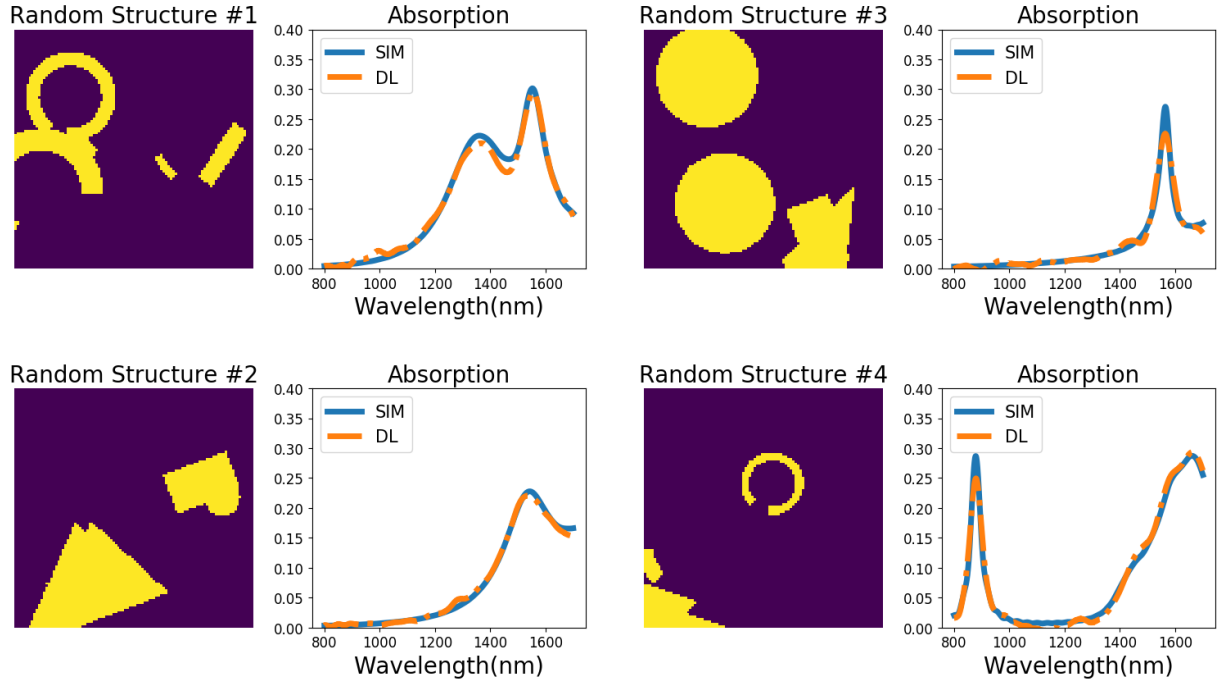

Fig. 4: The results of applying the final deep learning model on some random structures from the validation dataset. The model can almost predict the results completely. Random structures have many different properties and a good model should be able to predict all of them. The blue solid lines show the absorption curves obtained from Lumerical simulation and the orange dotted lines show the absorption curves predicted by the deep learning model.

## 3. Results

After finding the best model with the lowest loss on the test dataset (we defined the root mean square as the loss function) we fit the data on the validation dataset. The lowest achieved loss on the validation dataset was 4.2591e-05 after 500 iterations. The results of some of the structures from the validation dataset are shown in fig 4. As can be seen from this figure the model could almost predict the results exactly. To show what happens inside the deep learning model we provided the output of each layer in fig 5. For this figure we used a new structure to check the model one more time.

Since we had to run around 100,000 simulations we used a CPU server with 28 Intel Xeon CPU E5-2697 v3 2.6GHz cores. It took around 15 days to prepare these simulations. To implement the model we used Keras with the Tensorflow backend and the code was written in python. For implementation of CNN we used a GPU server with two GTX 1080ti graphic cards. It took around 3 days for the model to run. The same model takes much more on a CPU server.

One points is worth mentioning about getting these results. The reason that we used 100,000 simulations was that because of the complexity of the problem, we couldn't get an acceptable loss for test data with lower number of simulations, although we tried many different layouts for the model. So we had to increase the number of the input so the model can learn.

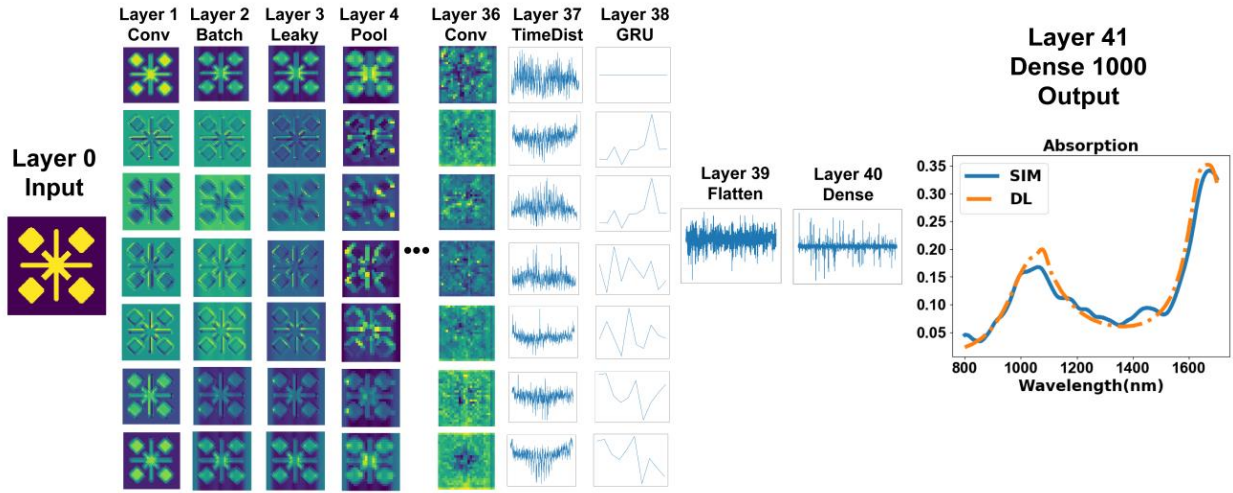

Fig 5: The output of some selected layers of the used deep model. We used a new structure different than the initial data to see how the model fits on it. The first layer is the input which is our structure and the last layer is the output which is the desired absorption curve. The model is consisted from 42 layers and each of these layers are consisted from different number of layers themselves. Since we couldn't show all the layers, some of the first and last layers are shown. For better comparison the final layer is combined with the expected results from simulation.

## 4. Generalization and Further Steps

The method introduced above can be easily generalized to cover other structures too. For example imagine that we wanted to consider other thicknesses too. We can run the same amount of simulations for the desired thicknesses like 50nm, 100nm, 150nm.... and then perform the model on all of these data. Now we have two inputs, the images and the thicknesses. One method for solving this problem is to add this extra factor in the fully connected layers. So first we extract spatial information with convolution layers and then add thickness parameter by concatenating its array with the first fully connected layer. This will lead to a model that can predict absorption of different structures having different thicknesses. Also different kinds of materials can be used by making the input images in color instead of black and white, where each color will be assigned to a particular material.

The above model can also be used as a discriminator for predicting structures with desired absorption curves too. This is the reverse of what we did here. This can be done by using generative adversarial networks (GANs). GANs are consisted form two neural networks, which one of them suggests a design, which is called the generator and the other one verifies it, which is called the discriminator. The generator starts from noise images and improve itself by getting help from the discriminator. Now if we have a good model for discriminator like the one that we introduced here we can predict structures with desired values[24, 25].

## 5. Conclusions

In here, we introduced a method that can predict optical properties of plasmonic structures by using deep learning. We first discussed how our structure can be converted into 2d images. We then used these images as the input of our neural network model. We talked about the model structure and how to improve the model. Finally we checked our model with the unseen data to verify it. We also discussed how this model can be generalized for other structures and how it can be used to predict structures for desired optical properties.